UJF-CNRS TIMC-IMAG Laboratory
IN3S – School of Medicine – Domaine de la Merci – 38706 La Tronche cedex – France
Jocelyne.Troccaz@imag.fr

**Abstract**
Medical robotics includes assistive devices used by the physician in order to make his/her diagnostic or therapeutic practices easier and more efficient. This chapter focuses on such systems. It introduces the general field of Computer-Assisted Medical Interventions, its aims, its different components and describes the place of robots in that context. The evolutions in terms of general design and control paradigms in the development of medical robots are presented and issues specific to that application domain are discussed. A view of existing systems, on-going developments and future trends is given. A case-study is detailed. Other types of robotic help in the medical environment (such as for assisting a handicapped person, for rehabilitation of a patient or for replacement of some damaged/suppressed limbs or organs) are out of the scope of this chapter.

**82.1. Introduction: clinical context and objectives**
Informatics and technology have dramatically transformed the clinical practice over the last decades. This technically oriented evolution was parallel to other specific evolutions of medicine:
- Diagnostic and therapy procedures tend to be less and less invasive for the patient aiming at reducing pain, post-operative complications, hospital stay, and recovery time. Minimal-invasiveness results in smaller targets reached through narrow access (natural or not) with no direct sensing (vision, touch) and limited degrees of freedom.
- More and more data are handled for each patient (e.g. images, signals) in order to prepare and to monitor the medical action and those multi-modality data have to be shared by several actors participating to this action.
- As in many domains, quality control gets more and more important and quantitative indicators have to be made available.
- Traceability becomes mandatory especially regarding the always increasing number of legal cases. Traceability is also of primary importance in cost management.

These evolutions make the medical action more and more complex for the clinician both at the technical level and at the organisational one. Computer-assisted medical interventions may contribute a lot to those clinical objectives. They may provide quantitative and rational collaborative access to patient information, fusion of multi-modal data and their exploitation for planning and execution of medical actions.

**82.2. Computer-assisted medical interventions**
The development of this area from the early eighties results from converging evolutions in medicine, physics, materials, electronics, informatics, robotics, etc. This field and related subfields are given several names almost synonymous: computer-assisted medical interventions (the most general), augmented surgery, computer-assisted surgery, image-guided surgery, medical robotics, surgical navigation, etc. We will subsequently use "computer-assisted medical interventions" or "CAMI" to name the domain.

Definition: CAMI aims at providing tools that allow the clinician to use multi-modality data in a rational and quantitative way in order to plan, to simulate and to accurately and safely execute mini-invasive medical interventions. Medical interventions include both diagnostic and therapeutic actions. Therapy may involve surgery, radiotherapy[1], local injection of drugs, interventional radiology, etc.

*82.2.1. CAMI major components*

CAMI [1] may be described as a perception-decision-action loop as presented in figure 1. The perception phase includes data acquisition and processing, the development of specific sensors and their calibration. Data may be acquired pre-operatively, intra-operatively or post-operatively. Images may provide anatomical information or functional one; data provided by the sensors may be 1D, 2D, 3D (sparse or dense), 4D (varying with time). Each imaging sensor brings its specific type of information and multi-modality is in general necessary. The need to assist the intervention in a quantitative and accurate way requires the calibration of sensors enabling both the transformation from image coordinates to spatial coordinates and the correction of possible image distortions. Specific sensors such as position sensors (also called localizers) and surface sensors have been integrated to CAMI components. Localizers give access to position and orientation of objects (instruments, other sensors, anatomical structures such as bones). Surface sensors give access to the external surface of an object (an organ for instance). All those information are potentially useful to plan and control the execution of a medical action.

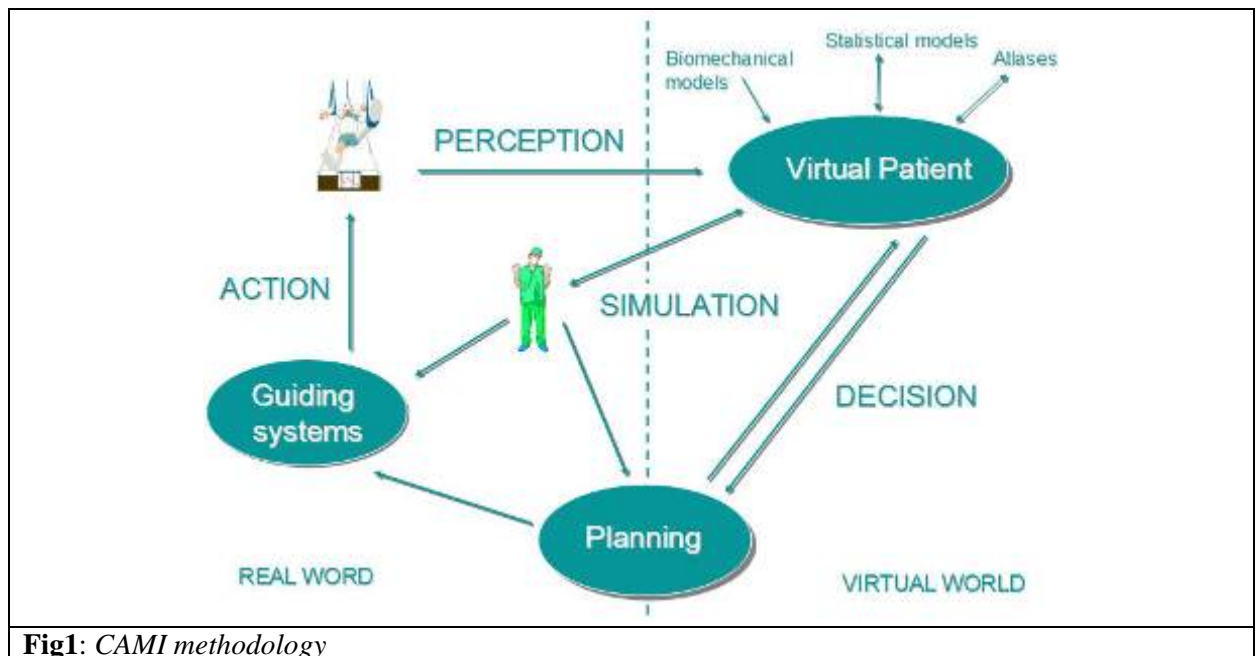

**Fig1**: *CAMI methodology*

The main objective of the decision stage consists in building an integrated numerical model of the patient and, in some cases, a model of the action. As mentioned previously, many types of information may be useful: data provided by imaging sensors for instance but also a priori medical knowledge (for instance, statistical data about organ shapes or occurrence of pathologies, biomechanical models of the limbs, etc.). One very important stage is data fusion, also called registration, which corresponds to the action of representing all the information in a single reference frame. Registration and in particular medical image registration [2], has been a very active domain for three decades. From this integrated model, the medical action can be planned: for instance, one may have to determine the type, size, position and orientation of a knee prosthesis that provide the best alignment of hip, knee and ankle joints; or one may decide which number, shape, intensity, position and orientation of radiation beams would allow to radiate a tumour of a given shape, with a given dose, whilst sparring organs at risks. Planning may involve very interactive tools where the clinician navigates in the data and specifies the selected strategy. It may also include optimization tools when the medical goal can be specified as an optimization problem (radiotherapy planning is an excellent example). In some cases, planning may be very difficult and simulators could provide help for the computation of the clinical outcome of a selected action: for instance, when a bone fragment has to be moved in the patient face, it may be very useful for the surgeon to foresee the functional and aesthetic consequences of this gesture on soft tissues [3]. Registration and planning are included in the decision stage.

---

[1] Some of the medical terms are explained in a glossary located at the end of this chapter; such terms are written in blue the first time they are used in this text.

Action requires the accurate execution of the planned intervention. Often, a registration stage is necessary to transfer the planned action to the interventional conditions; for instance, the action is planned from pre-operative data (e.g. MRI) and must be registered to the real patient in intra-operative conditions. Two main types of assistance exist: navigational aids and robots. In the first case, the action is monitored using suitable sensors such as localizers; information is rendered to the clinician about the planned and executed actions. The clinician makes use of this information to control his/her action. As presented in chapter 8.3, such systems are called "passive". The first surgical navigators have been used for neurosurgery [4,5]. Chapter 8.3 gives some details about such navigators. The action can also be performed more of less autonomously by a robot. The first application of a robot in CAMI took place around 1985; the application field was also neurosurgery. The medical robot needs to be connected to patient data and models in order to be able to transfer the planned intervention in the robot coordinates; this is the robot registration problem. The medical robot is therefore always an *image-guided robot*.

Simulators may also be developed for training purposes. The advantages are the abilities to provide frequent and rare medical cases, to gain realistic experience with lower stress than with real patients and to quantitatively evaluate the practitioner. This may facilitate the acquisition of new medical skills, with new techniques and/or tools. This can also be considered as a very valuable component of CAMI.

Based on numerical data, tracking of objects (instruments, sensors and anatomical structures) and positioning of tools with navigation or robotized aids, those CAMI procedures are fully traceable.

*82.2.2. Added value of a robot*
Automation is generally not a primary goal of medical robotics where the interaction with a clinical operator has to be considered with a very special attention. Indeed, most often medical robots are not intended to replace the operator but rather to assist him/her where his/her capabilities are limited. In general CAMI systems are considered only as evolved tools in the hands of the clinician.

In medicine, like in many other application areas, the robot advantages are in its precision, ability to repeat a task endlessly, potential connection to computerized data and sensors, capability to operate in hostile environments (biological or nuclear contaminations, war or catastrophe areas, space (orbital station) or undersea (submarine), etc.) where clinicians' presence or abilities may be limited and persons may need medical care.

Navigational aids have already demonstrated their clinical added value in various specialties (neurosurgery, orthopaedics in particular) and their integration in the clinical environment is generally easier than for a robot. Safety issues are also more limited. Moreover navigation systems are very often more cost-effective. Those are the reasons why it is very important to use a robot only for clinical applications where it can offer functionalities that the navigation system cannot; potential specific robot abilities are:
- To realize complex geometric tasks (for instance to machine a 3D bone cavity);
- To handle heavy tools (e.g. radiation apparatus) or sensors (e.g. intra-operative surgical microscope);
- To provide a third hand to the clinician;
- To be remotely controllable and to offer scaling capabilities in terms of transmitted motions or forces;
- To filter undesired movements (such as physiological shaking in tele-operation);
- To be force-controllable down to very small force scales;
- To execute high resolution, high accuracy motions (for microsurgery);
- To track moving organs and to be synchronized to external events based on some signals;

- To be introduced in the patient for intra-body actions.

Another aspect that will be further discussed is the absolute necessity to demonstrate the clinical added value of the system i.e. to prove that it brings a clear clinical benefit at some level (for the patient / for the hospital / for the healthcare system).

**82.3. Main periods of medical robot development**
*82.3.1. The era of "automation" (1985-1995[2])*
Early medical applications of robotics are characterized by transferring to this domain accurate and automated tool positioning capabilities of the robot originally developed in industrial applications.

As mentioned previously the first surgical robots were introduced in neurosurgery. Neurosurgery had already a very long tradition of minimal invasiveness and use of computerized 3D imaging data; indeed, the first CT scans in the early 1970's were performed for brain imaging. Stereotactic neurosurgery is a particular type of minimally invasive procedure which consists in "blindly" introducing a linear tool in the brain through a keyhole (3mm diameter) for biopsies, removal of cysts or haematomas, placement of stimulation (Parkinson disease) or measurement (epilepsy) electrodes, etc. Conventionally, stereotactic neurosurgery is performed with the help of a stereotactic frame which functions are to immobilize the patient skull, to register pre-operative data to the intra-operative situation – the frame is installed on the patient's head before the pre-operative exam – and finally to offer mechanical guidance for tool insertion. Except for the first function, a robot can advantageously replace the frame: it is easily connected to imaging data, is less invasive and may offer a larger range of trajectories for tool positioning. Though, anthropomorphic robots associated to stereotactic frames or robotized stereotactic frames have been developed and clinically evaluated in the early eighties (for instance [6,7]). [7] describes the accurate positioning of a guiding tool with respect to pre-operative data in a stereotactic neurosurgery application using a Puma 260 robot. The cited paper reports a series of 22 patients. [8] presents a CAMI system integrating a modified industrial robot (reduced speed, non backdrivable joints) for a similar application. The first patient (see figure 2a) was operated with this system in 1989 and since hundreds of patients have been treated with this technology. This system was the academic precursor of the Neuromate product (see figure 2b) with which thousands of patients were treated. Those systems are called "semi-active" because the robot is only a guiding device and the gesture (drilling the skull and inserting the needle) is still performed by the surgeon through a mechanical guide positioned by the robot.

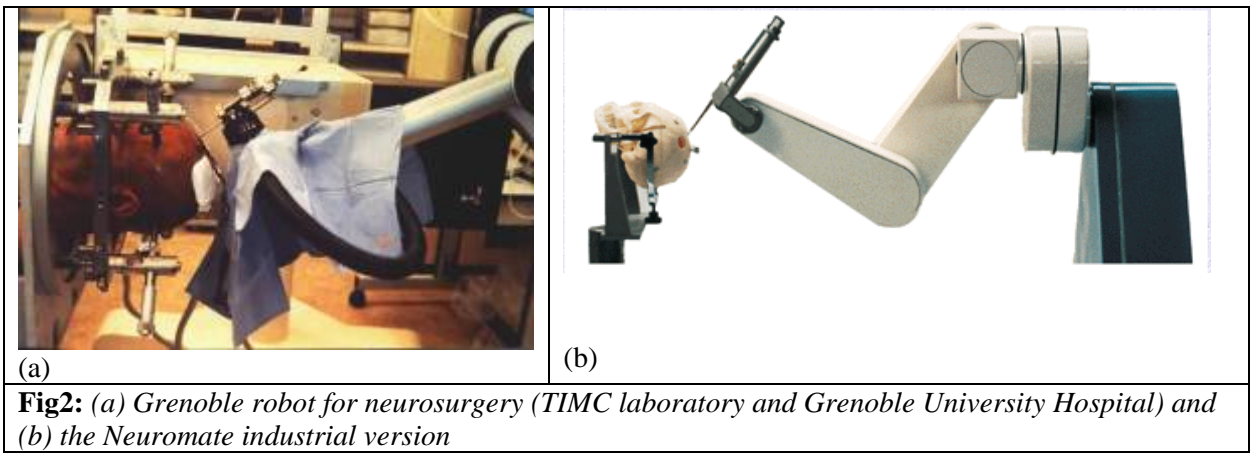

(a) (b)
**Fig2:** *(a) Grenoble robot for neurosurgery (TIMC laboratory and Grenoble University Hospital) and (b) the Neuromate industrial version*

[9] proposed an automated view of the whole process of stereotactic neurosurgery: a robot equipped with different types of tools in a tool-feeder effector was installed in a CT room; after imaging data and planning the robot performed the drilling of the bone and the placement of the surgical tool. CT enabled repeated image control. The robot was specifically designed for this application. The first two

---

[2] Dates do not strictly define a period but rather provide a general notion of activity.

patients were operated in 1993. [9] reports eight cases of biopsies. As far as we know, this system has not been extensively used in clinical routine.

In orthopaedics, the placement of prostheses requires the preparation of the bones: for instance a cavity has to be drilled before inserting the femoral component of a hip prosthesis; similarly, planar cuts have to be realized on the tibia and femur extremities in order to place knee prosthesis components. Thus, such stages of the interventions are very close to machining a mechanical part: a 3D shape has to be accurately realized by sawing or milling with given position and orientation. This is why the idea of using a robot came very naturally for such tasks. Robodoc [10,11] used for cavity preparation in total hip arthroplasty was developed from 1986 first in a laboratory set-up. Then, from 1989 to 1991, 29 dogs were operated with the help of the system. The first 10 patients were included in a FDA-approved clinical research trial between 1991 and 1993; very large series of patients for comparison of traditional interventions to Robodoc-assisted ones were included from 1995 to the early 2000's. Thousands of patients benefited from the use of Robodoc. Robodoc is "active" in the sense that a part of the surgical action (machining the bone) is performed autonomously by the robot under surgeon supervision. Several other systems were developed based on a similar approach.

The underlying idea in this first period was that surgical subtasks such as accurate positioning of a tool or machining, based on numerical data, could be transferred to a robot and automated to a certain extent.

*82.3.2 The era of interactive devices (1990-2005)*
Whilst the automation era focused on rigid and non deformable anatomical structures, the second era is characterized by the development of more interactive control schemes for complex tasks in particular for interventions on soft tissues.

Indeed, in the two previous examples (cf. section 8.8.3.1), the anatomical structure of interest is rigid and non deformable: in the case of stereotactic neurosurgery, the brain is accessed though a small hole. The trajectory through the hole is simple (a linear tool is inserted to a given target). The brain motion and deformation can be neglected and the skull is immobilized. As regards orthopaedics, bones are not deformable and can be fixed to external fixtures to avoid any motion. Obviously there are much more clinical situations where the procedures are more complex and concern soft, mobile and deformable tissues. For such applications automation is often still out of reach or may not be the preferred solution; for instance when the expertise of the clinician is so complex that it cannot totally be transferred to the robot.

In contrast to earlier automated robot control, robotic development in the mid 1990s was characterized by more direct operator control. In particular efforts were put towards tele-operation; this form of robotic application was traditionally used in the nuclear industry. In this situation, the surgeon is totally in control of the surgical tool through a master-slave apparatus. The distance of the master and slave components is highly variable: although the main use is for very close tele-operation (in the same room), some very long distance (thousands of km) experiments have taken place. The function transferring movements from the master to the slave may involve scaling down (forces and/or motions) and filtering. The main clinical applications are in endoscopic surgery where instruments and optics are introduced in the patient's body through small incisions. Those entry points limit the instruments possible motions (4 degrees of freedom (dof) instead of 6) and the surgeon has to operate under video-control. In a first stage, the motion of the optical system – the endoscope – alone was robotized: instead of requiring one assistant to move the endoscope for a hand-eye coordination, the robot is controlled by the surgeon himself/herself by different means: voice control (AESOP [12]), head movements (Endoassist [13]) or high level image processing software (see 8.8.4). More recently the displacement of instruments has also transferred to robots, potentially offering extra-dofs; this is the case of the DaVinci multi-arm system that offers intra-body dofs of the instruments [14].

From the mid nineties synergistic devices [15] also named "hands-on" robots [16] were proposed. The rationale for such systems is that the clinical application is generally so complex that it can only be partially embedded in numerical models and data. Therefore relying both on the clinician for his/her very high skills, capacity of judgement, intelligent perception and on the computerized robot for its quantitative knowledge of the planning, accuracy and sensors is potentially very fruitful. In the Padyc [17,18], Acrobot [16, 19] or Makoplasty [20] systems, the surgical tool is attached to the robot effector and the surgeon holds it. The motions proposed by the human operator are filtered by the robot in order to keep only the part of the motions which are compatible with the surgical plan: for instance, the tool has to keep in a plane or in a given region. Different technologies implement this principle of constrained motions: clutchable freewheels, backdrivable motors, controlled brakes, etc.

*82.3.3 The era of small and light dedicated devices (2000-now)*
This era tends toward a miniaturization of robots up to the stage where they can be attached to a body part, inserted into the patient's body or integrated to mechatronic surgical instruments.

In systems presented in sections 8.8.3.1 and 8.8.3.2, the robot mechanical architectures were very inspired by traditional industrial robotics: general purpose anthropomorphic 6 (or more) dofs arms are used. Systems are potentially very versatile and can adapt to a variety of tasks. But this generality has a cost: these robots are also very often quite cumbersome and are attached to the floor, the bed or to the ceiling. Their working space may be quite large and not easily manageable in an operating room (OR) that offers limited space and in which people (patient and staff) are very close. This is why new generations of robots have been specifically designed and developed for limited application areas.

This new generation of systems has produced robots so small that they can be supported by the patient body or attached to one of his/her bones. Because of their size these robots are more easily integrated to the OR and they are designed to be very well-suited to a specific task. These two factors offer potential advantages over traditional systems by providing a safer operating environment and cost-effectiveness. The Grenoble TIMC laboratory was pioneer in small scale robotics and figure 3 illustrates four such pioneering works. TER [21] (cf. figure 3a), a robot for tele-ultrasonic examination, has been clinically validated for remote examination of abdominal aortic aneurysm and for emergency care of abdominal traumas; LER [22], a light endoscope holder, has been validated on pigs and corpses and VIKY, the corresponding industrial product (cf. figure 3b) is EC marked and has been used on patients; LPR [23] (cf. figure 3c), a CT/MRI compatible robot for punctures, has been validated for CT on pigs and for MRI on volunteers (without puncture!); Praxiteles [24] (cf. figure 3d), a robot for total knee arthroplasty, has been validated on corpses and recently entered a multi-centres clinical evaluation on patients.

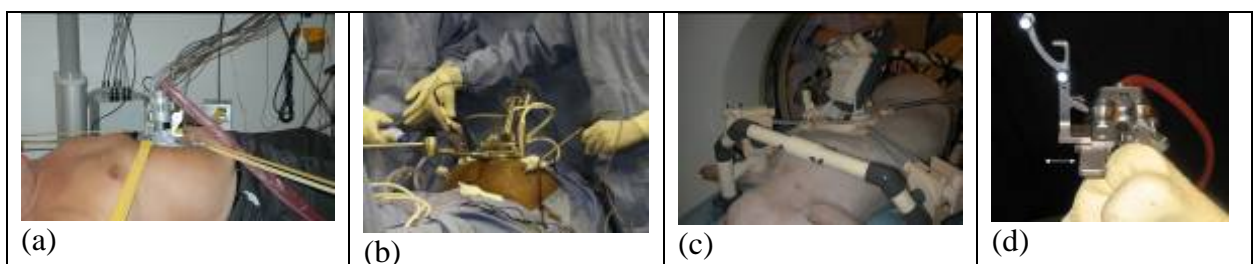

**Fig3:** *On-body and on-bone robots at Grenoble : (a) TER slave robot for tele-ultrasound examination (TIMC, Grenoble University Hospital), (b) LER-Viky endoscope holder (TIMC, Endocontrol-Medical company, Grenoble University Hospital and Paris La Pitié Salpétrière Hospital), (c) LPR MRI and CT-compatible robot for image-guided punctures (TIMC, Grenoble University Hospital), (d) Praxiteles robot to total knee arthroplasty (PRAXIM, TIMC, Brest University Hospital)*

Following the same philosophy, the Mars system now called Mazor [25] – the product – is presented in details in chapter 8.3. It is being clinically evaluated on patients for spine surgery. This list of "body-supported" or "bone-mounted" devices is not exhaustive.

Other systems are sufficiently small to be completely introduced in the body of the patient. Some groups focus on locomotion issues. For instance, [26] has proposed several versions of an inchworm type of compact robot that moves inside the intestine. The aim is to replace the long, quite rigid and painful endoscope traditionally used in colonoscopy; experiments on pigs have been carried out successfully. [27] develops, with a rather similar locomotion principle, a robot that moves on the heart surface; it has also been experimented on living animals. [28] describes a robot rolling on soft tissue organs in the abdomen; this robot carries a camera. It has also been experimented on animal organs. Finally, [29] proposes a smaller and simpler device which is actuated from the outside of the body using external magnetic fields; such a robot would be injected in the eye for instance for drug delivery in the retina vessels.

Other groups aim at giving extra dofs to traditional instruments: the domain of active catheters that can adapt actively to vessel curvature has been investigated for several years [30,31]. Recently, some of these systems have reached the market (see table 1 in section 8.8.7). Finally, the field of articulated tools for endoscopic surgery is also very active; the aim is to develop surgical tools equipped with intra-body dofs to recover full mobility of the tools with respect to the organs. The specific area of intra-body devices raises very challenging issues related to biocompatibility, safety, power supply and data transmission. Smart pills such as the M2A [32] or Norika [33] ones, that are swallowed by a patient and allow visualizing the gastro-intestinal track do not integrate yet action devices but can be seen as precursors of those future highly integrated mechatronic intra-body devices.

## 82.4. Evolution of control schemes

Evolution from "bone applications" to soft tissues also resulted in an evolution of control schemes giving a more active role to the surgeon or aiming real-time perception-decision-action control loop.

A large number of the oldest systems integrate a "single shot" perception-decision-action process. For instance a planned trajectory is selected from CT data and is transferred to the intra-operative conditions after registration of pre-operative data to intra-operative ones. To guarantee that the plan is still valid intra-operatively, the anatomical structure of interest must not move. This approach has been largely used for neurosurgery and orthopaedics surgery; in both cases the structure is fixed using a stereotactic frame or external fixtures. The operator is often limited to a role of supervisor when the robot moves the tool.

More recently (cf. 8.8.3.2) the operator has been given a more active role in the execution of the task. In the case of "co-manipulation" the robot and the operator participate simultaneously to the motions of the tool. In the Acrobot system [19] the PID coefficients of the control law are given by functions that depend on the position of the tool with respect to a region of allowed motions – the region corresponds for instance to the bone to be removed for prosthesis implantation : in the inside of the region, the motions proposed by the operator and detected by a force sensor are transmitted to the tool without significant modification; in an intermediate region the motions are more or less transmitted depending on their direction; motions outside the planned region are strictly forbidden. The Steady hand [34] works similarly: the operator manipulates a handle and proposed motions are detected. In contrast to the Acrobot, the Steady Hand does not select permitted directions of motions but it scales down motions and forces for microsurgery or biology applications. In the case of Padyc the principle is a little different [17,18]; Padyc is a passive arm for which each joint is equipped with two freewheels than can be independently clutched or unclutched using a motor; the motor is velocity-controlled and this velocity determines the range of allowed motion of each joint in each direction at each instant. This range of motion is computed from the representation of the task (position to reach, trajectory to follow or region to keep inside for instance) and from the current position of the robot. No force sensor is necessary. For tele-operation, the operator interacts with a master device and the slave robot reproduces the motions proposed by the operator. Transfer functions may enable scaling and filtering of disturbing shaking motions like in the DaVinci. Tele-operation can integrate a force feedback to the operator like in the Sensei system for endocardial robotized catheter control.

Few systems integrate hybrid force-position control of the robot. One application case is external ultrasound examination which requires a constant contact of the ultrasonic probe on the body of the patient. Thus several robotic systems for ultrasound examination (remote or automated) were developed with a hybrid control scheme [35,36]. They generally combine position control in the main direction of motion with force control in order to keep the contact during probe motion. Dermarob [37] developed for skin sampling for skin graft in burnt patients follows a similar hybrid control scheme.

Some of the systems developed for orthopaedics and neurosurgery offer some basic tracking abilities using a localizer and markers attached to rigid structures of interest. This is for instance the case of Caspar used for ligamentoplasty in knee surgery and the case of a frameless version of the Neuromate. Systems like Mazor or Praxiteles which are mounted on the anatomical structure of interest move with it and therefore suppress the tracking problem. Recently other systems have been developed with evolved abilities for tracking of soft tissues. Instead of having a fixed target, the robot has to track it in real time in general from information extracted from imaging data. Typical examples are related to organ motions induced by cardiac or respiratory cycles. In those cases, a real-time perception-decision-action loop must be developed. The Cyberknife system (cf. section 8.8.5) developed for radiotherapy [38,39] determines the current position of a tumour moving with patient respiration by using external markers localized in real time [40,41]. Alternate approaches consist in coupling motion in the image information to motion of the robot at a low level through visual servoing. Pioneer work was described in [42]; from high speed camera data, the motion of relevant fiducials on the heart surface was computed and fed back to a slave robot for tele-operation giving the operator the feeling of a stable target. More recently several groups have applied visual servoing to image-guided minimally invasive actions. Video images are used for robotized endoscopy [43,44]; ultrasonic images have also been introduced [45,46] for cardiac and vascular applications. Directly coupling motion detection in the images to motion of the joints requires very careful robustness analysis.

**82.5. A case study: the Cyberknife® system**
This system is described in more details to make a little more visible the intrinsic complexity of a robotic CAMI system. This section is also intended to introduce the potentially long road from an initial paradigm and a first prototype to a clinically used product. Specific issues related to medical robotics and explaining this long road are discussed in the next section (8.8.6).

The Cyberknife system distributed by Accuray Inc. has been developed in the context of radiotherapy. Usually, when treating a patient with conventional linear accelerators, the patient is positioned onto a couch that has 4 main dofs (3 translations and 1 rotation) and the linear accelerator allows orienting the radiation beam using two more rotational dofs. The convergence point of the radiation beams is named isocenter. A reference frame $R_{iso}$ centred on this point is associated to the complete radiation system. The dofs enable positioning the tumour on the isocenter and orienting the radiation beams properly with respect to the tumour in order to execute the planned treatment (see 8.8.2.1). In practice, due to the complexity of positioning the patient and orienting the beams with such machines, treatments generally consists of rather simple ballistics with rather few radiation beams. The Cyberknife concept proposes to install the radiation source on a 6 dofs robot in order to avoid repeated combined motions of the couch and of the linear accelerator. This allows executing complex treatments with hundreds of beams distributed around the tumour without having to move the patient. Using such a large number of small beams enables sculpting the dose distribution to the tumour shape with high accuracy.

The first version of the Cyberknife [38] was installed at Stanford Medical Centre in February 1994 and the first patient was treated with the system on June, 8[th], 1994. The current version of the system integrates a Kuka robot (see figure 4a). In early May 2008, the Accuray Company communicates 134 installed systems worldwide and 40000 patients treated mostly for brain, spine, lung, prostate, liver or pancreas tumours.

As regards patient data, pre-operative CT is traditionally used for treatment planning; other modalities can be fused to CT data such as MRI or PET for planning refinement. Intra-operative X-ray imaging produced by two X-ray systems enables patient initial positioning and participates to tracking. The system is calibrated which means that the spatial relationships between the different reference frames

$R_{iso}$, $R_{robot}$ and $R_{X-ray}$ (associated respectively to the isocenter, robot and X-ray devices) is determined using specific procedures and calibration objects. Such calibration is necessary to transform intra-treatment imaging information into robot positions with respect to the radiation system. The transfer of planning information from $R_{CT}$ to $R_{iso}$ is possible thanks to image registration procedures more detailed in the following sections.

In the first version of the system dedicated to neuro applications, after an initial set-up, patient motion was detected and corrected before each dose delivery of a single beam. Thanks to the CT exam, two synthetic X-ray images called DRRs (Digitally Reconstructed Radiographs) are computed. They correspond to what would be seen by the two X-rays machines in the treatment room for a perfect position/orientation of the patient. Two real images are acquired just before dose delivery and automatically compared to the DRRs; this results in the computation of the patient shift from the ideal position/orientation. For small errors the robot position/orientation with respect to the patient is automatically corrected by combining this information with calibration data. For larger discrepancies, a re-planning phase is necessary to avoid any collision during robot repositioning.

A later version of the system enables real-time tracking of organs that move with patient respiration (lung, liver, kidney for instance) [40]. As regards patient respiration during treatment several approaches are used in conventional radiotherapy: the less accurate method consists in enlarging the targeted zone to account for tumour motion; irradiating healthy tissues is more acceptable than missing cancerous cells. This enlarged zone can be computed by combining information from two scanners acquired at the end of the exhale and inhale phases. A second approach consists in synchronizing dose delivery to a given stage of the respiratory cycle (end of inhale for instance) but this stage has to reasonably repeatable and has to be detected reliably. In a more sophisticated approach, such as developed in the Synchrony® version of Cyberknife, the system tracks the organ motion and the robot follows this motion during dose delivery in order to execute properly the planned treatment [41]. Since it is not possible to get the tracking information from X-ray images – such organs may not be visible on the radiographs and continuous imaging would result in over-irradiating the patient – a localizer is introduced in the treatment room. This localizer associated to $R_{loc}$ and calibrated with respect to $R_{iso}$ is used to track passive markers placed on the patient chest; it delivers the marker positions about 20 times per second. X-ray images can be acquired about every 10 seconds. Because the motion of internal organs is different from chest external motion internal radiopaque markers are implanted close to the tumour before the pre-operative CT exam. The relationship between the tumour position and the internal markers is determined using the CT data; the internal markers enable the initial patient set-up by registration of CT-data to X-ray images. The relationship between internal markers visible on X-ray images and external markers tracked by the localizer is learnt in a preliminary stage where both data are acquired during several breathing cycles. During dose delivery, between two successive X-ray acquisitions, the position of internal markers is interpolated from the position of the external markers and from the model; each new acquisition enriches the model by adding a new couple of synchronous positions of internal and external markers. Finally, the position of the tumour can be deduced from the internal markers and fed back to the robot for tracking and accurate dose delivery in spite of organ motion. The different components of the tracking function are illustrated on figure 4b. Current work deals with non invasive alternatives to the implanted internal markers.

As can be seen in this example, a lot of hardware and software components are associated to the robot and many stages are involved in the real-time determination of the robot position with respect to the patient. To provide the required accuracy for precise dose delivery, errors have to be reduced as far as possible for each of these elements; this is specially demanding. Moreover, because the robot develops very large forces and torques in the immediate vicinity of the patient, reliability and robustness are mandatory. Unfortunately, as for many other commercialized systems, due to intellectual property issues few data are available concerning detailed industrial developments and technical testing.

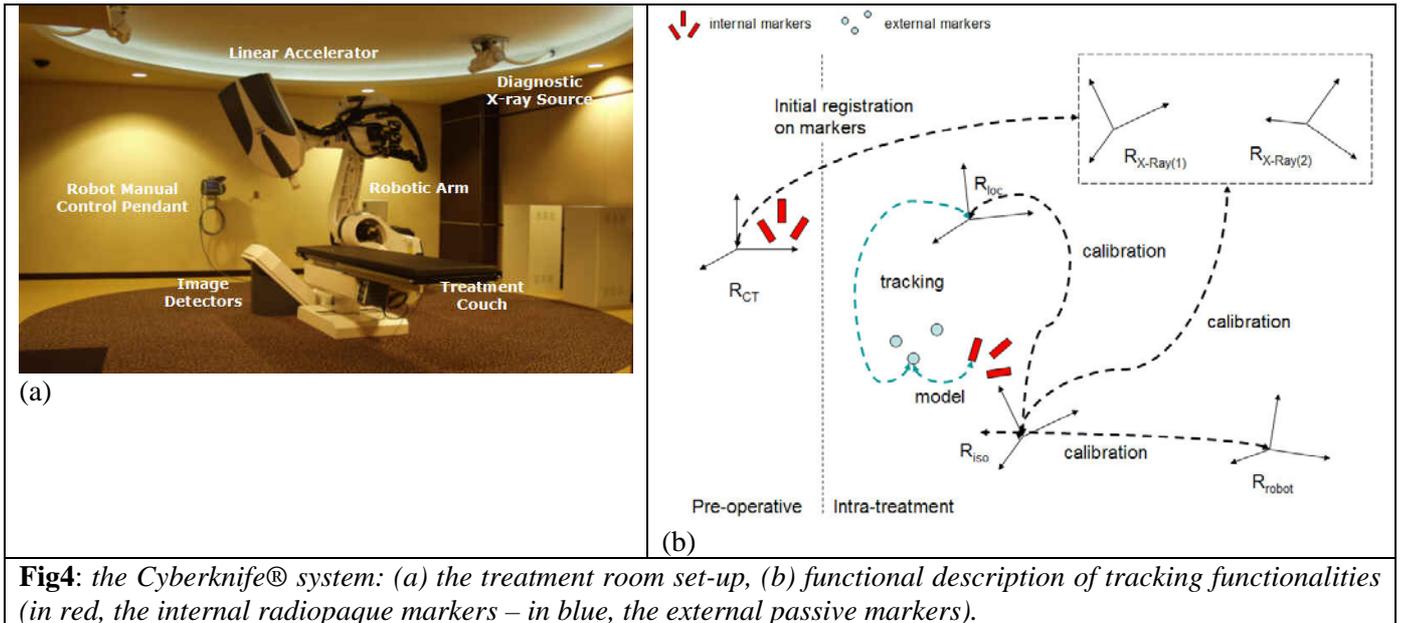

**Fig4**: *the Cyberknife® system: (a) the treatment room set-up, (b) functional description of tracking functionalities (in red, the internal radiopaque markers – in blue, the external passive markers).*

The Cyberknife approach supposes the ability to install a linear accelerator as end-effector of the robot; the Cyberknife robot typically carries 6MeV accelerators which are quite compact and light as compared to other radiation devices. When higher power is necessary or when radiation beams are produced by synchrotrons and/or cyclotrons – for proton therapy for instance – the radiation source cannot be positioned and oriented by a robot. Alternative systems (e.g. [47]) have been proposed to robotically position the patient relative to the radiation beam by means of parallel robotized seats or robotized couches.

**82.6. Specific issues in medical robotics**
As it has been briefly introduced in the previous sections, medical robotics raises specific issues at the technical, clinical and organizational levels in a very intricate way. This section discusses those issues in more details.

Firstly this type of robot has to be used in a human environment. Because these robotic systems generally require a close collaboration with a clinician, specific man-machine interface questions have to be resolved. In case of surgery, the operator cannot interact easily with classical man-machine interfaces because he/she has to work under sterile conditions. This motivates the development of specific interfaces such as voice control or foot pedals. In case of co-manipulation, the part of the robot that is held by the clinician must be made sterile before each intervention. Because the robot is used in close proximity to human beings – at least the patient – safety issues are mandatory. Safety has to be demonstrated both at the hardware and software levels. Different approaches are possible [48]. The choice of specific architectures of robots or co-manipulation control modes may solve part of the problem. Norms and regulations are intended to guarantee that the system is safe. However, it would be certainly interesting to develop such systems using specific design methodologies introduced in critical applications (aeronautics, nuclear plants, etc.) to really anticipate the behaviour of such a complex system and to avoid any misuse of it. This direction is still to be explored.

The clinical use of the robot also requires electro-magnetic compatibility with the environment in particular in the OR. When the robot is used in specific environments such as inside a CT or MRI imaging sensor, the robot must not disturb the image acquisition or corrupt the data. This may significantly constrain design choices and selected materials: for instance in the MRI environment no ferro-magnetic part should be integrated to the robot. Sterile cleaning of the robot is also an issue. Bone-mounted robots which are in very close contact to the patient must be completely cleanable. This is the case for instance of the Praxiteles and LER-Viky systems (cf. 8.8.3.3) which are autoclavable. In the case of bigger robots, their end-effector may be introduced inside sterile disposable plastic bags.

The robot connection with the tool which is in contact with the patient has however to be cleaned in a sterile way. Finally, the system and the robot must be designed such that the robotic procedure can be easily and rapidly converted into the conventional intervention, at any time, in case of problems. This may also have strong design consequences.

As mentioned previously, introducing a new medical device requires demonstrating the added value over existing techniques from a clinical perspective. For instance, using the robot could enable performing less invasive interventions resulting in shorter stay of the patient in the hospital. In the same way, using a robot to hold the endoscope may save one assistant who can be transferred where his/her skills are used in a much better way. But obviously evaluating such organizational benefits requires a very precise cost and resource management: all necessary human or material resources participating to the intervention from its diagnosis to its long-term follow-up have to be taken into account. On a more medical level, using a robot to machine the bone for prosthesis placement may may result in longer life duration of the prosthesis and/or less joint disjunctions (for hip for instance). In a similar way using a complex radiotherapy treatment delivered using a robot may allow dose escalation and may result both in a better control of a tumour and in fewer complications. In general, it is not easy to predict how accurate and sophisticated a robotic procedure should be to make a significant difference on the clinical level. To prove it may also be very challenging. The evaluation of clinical benefits may require long term trials involving several centres and many patients. Those clinical trials have to be conducted in accordance with the ethical standards and regulations of the concerned country; those standards and regulations may be very strict but they vary from country to country. Finally, added value may also be evaluated in terms of commercial advantage for the hospital that may attract more patients when high-technology is used for painless, minimally invasive procedures. The added value should be significant enough to compensate drawbacks related to the introduction of a robot such as for example the increase of procedure duration which is often observed even when the learning period is finished.

Cost is obviously another issue; indeed, several of the distributed systems are quite expensive. If we consider some the systems (for instance Robodoc, Caspar, DaVinci, Cyberknife) listed in table 1 of next section, costs average between one and two million USD. Aesop was around one hundred thousand USD. Frequently a 10% maintenance over cost per year has to be added and some of these systems generate an extra-cost per intervention (for instance about one to two thousand USD for the DaVinci). This may be quite a heavy investment for hospitals and clinics. Moreover, depending on healthcare funding models in the different countries, some of those costs may not be affordable by health insurances. The higher the investment, the more significant the added value has to be to justify the expense. More recently developed systems (smaller and simpler robots, disposable devices) are likely to propose more affordable solutions.

### 82.7. Systems used in clinical practice

For twenty years many medical robotic systems have been developed in laboratories and have been evaluated to a certain extent. Evaluation is twofold: at the technical level it consists in characterizing accuracy, reliability, robustness, etc. This stage may be realized on laboratory set-ups using phantoms that mimic more or less realistically the concerned part of the body. At the clinical level, experiments with corpses or animals enable a first approach to a more realistic evaluation of clinical feasibility and performances. Finally, a study on series of patients is always necessary to fully evaluate the system and its clinical added value. Relatively few medical robots underwent the whole evaluation process, have reached the market and have been largely clinically used. There are indeed two major challenges: how to turn a laboratory prototype into a certified product? How to make this product an industrial success? The reasons for this still limited diffusion of medical robotics in the clinical world certainly come from the specific constraints of medical robotics listed above and probably from the questionable added value of the robot in a number of cases. The complexity of clinical evaluation, certification and marketing also makes the process very long and expensive: for instance, in orthopaedics demonstrating the advantage of robots over competing techniques may last more than 10 years since the stability and life duration of prostheses cannot be demonstrated any earlier. In the same time, to evaluate them it is necessary to install robots in hospitals, sometimes at the company expense.

Convincing a hospital to buy such expensive devices before any medical evidence of the added value is particularly challenging.

Table 1 attempts to list as largely as possible the industrial systems that are or have been significantly clinically used in routine and emerging products; this table is intended to give a flavour of the clinical spreading of the technique. Numbers of systems are estimates established in early 2008. As can be seen several systems are no longer distributed: this deserves further comments. Surgiscope and MKM which are both "surgical microscope holders" probably faced a limited "added value versus cost" ratio. As regards Robodoc, this system had not yet demonstrated a clinical benefit when a misuse of the robot resulted in many clinical complications and legal cases in Germany; this ended up with the removal of the robot from the US and European markets. Caspar which offered functions very similar to Robodoc ones probably suffered from the same unproved added value and from the Robodoc "failure". Aesop and Zeus are no longer distributed due to Intellectual Property conflicts. Aesop was however very successful in clinics and could be certainly considered as an industrial success.

| System name | Clinical specialty | Type | Last distributing company | Estimated number of installed systems | Status |
|---|---|---|---|---|---|
| Neuromate | Stereotactic neurosurgery | semi-active (mechanical guide) | Schaerer-Mayfield | 15 < nb < 20 | unknown |
| PathFinder | Stereotactic neurosurgery | semi-active (mechanical guide) | Prosurgics Ltd. | unknown | unknown |
| Surgiscope | Microsurgery | surgical microscope holder | ISIS | 15 < nb < 20 | no longer distributed |
| MKM | Microsurgery | surgical microscope holder | Carl Zeiss | unknown | no longer distributed |
| Robodoc | Orthopaedics (knee, hip) | automated machining of bones | ISS Inc. | 70 < nb < 80 | no longer distributed in the US and in Europe |
| Caspar | Orthopaedics (knee, hip) | automated machining of bones | URS ortho | 50 < nb < 60 | no longer distributed |
| Cyberknife | Radiotherapy | Positioning and motion of the radiation device | Accuray Inc. | 134[3] < nb | growing |
| DaVinci | Endoscopic procedures (cardiac, digestive, gynecologic, urology, etc.) | endoscope and instrument holder - intra-body dofs | Intuitive surgical Inc. | 875[4] < nb | growing |
| Zeus | Endoscopic procedures | endoscope and instrument holder | Intuitive surgical Inc. | unknown | no longer distributed |
| Aesop | Endoscopic procedures | endoscope holder | Intuitive surgical Inc. | 800 < nb < 1000 | no longer distributed |
| EndoAssist | Endoscopic procedures | endoscope holder | Prosurgics Ltd. | unknown | Unknown |
| Naviot | Endoscopic procedures | endoscope holder | Hitachi | unknown | Unknown |
| Lapman | Endoscopic procedures | endoscope holder | Medsys | unknown | Unknown |
| Viky | Endoscopic procedures | on-body endoscope holder | Endocontrol Medical | probably < 10 | Emerging |
| Acrobot Sculptor | Orthopaedics (knee, spine, etc.) | "hands-on" robot | Acrobot Ltd | unknown | Emerging |
| PIGalileo CAS | Orthopaedics (knee) | on-bone semi-active | PLUS Orthopedics AG | unknown | Unknown |
| Praxiteles | Orthopaedics (knee) | on-bone semi-active | Praxim | probably < 10 | Emerging |
| Makoplasty | Orthopaedics (knee) | "hands-on" robot | Mako Inc. | probably < 10 | Emerging |
| Mazor | Orthopaedics (spine, etc.) | on-bone semi-active | Mazor Surgical Technologies | 10 < nb < 20 | Emerging |
| Estele | Radiology (ultrasounds) | tele-robotics | Robosoft | probably < 10 | Emerging |

---

[3] Number of installed systems in early May 2008 (source Accuray Inc.)
[4] Number of installed systems in early May 2008 (source Intuitive Surgical Inc.)

| | | | | | |
|---|---|---|---|---|---|
| Sensei | Interventional radiology (cardiology) | tele-operated robotic catheter for heart mapping (intra-cardiac) with force feedback | Hansen Medical | unknown | Emerging |
| CorPath | Interventional radiology (cardiology) | tele-operated catheter (intra-coronary) | Corindus | on evaluation - not yet distributed | Emerging |

**Table 1:** *Industrial medical robots*

The success of DaVinci is probably in its ability to offer intra-body dofs for endoscopic surgery; laparoscopic radical prostatectomy is one of the main clinical vectors of the large dissemination of the DaVinci system. As regards Cyberknife, the ability to perform complex dose distributions with tenths of small radiation beams and the capacity of radiating with accuracy the tumour during patient respiration are probably the keys of this success. Moreover, conventional radiation apparatus are expensive devices and cost issues may probably be less critical in the case of radiation therapy than for surgery.

This table also shows that emerging systems are often based on quite different design philosophy: small dedicated systems, more interactive control schemes. As can be seen favourite applications are in endoscopy and in knee arthroplasty where the market is large (6,000,000 of laparoscopic surgeries per year worldwide and 600,000 knee prostheses per year worldwide).

Many other systems are somewhere in between the academic and industrial worlds; there are not mentioned in this list. Several of those forthcoming devices concern the introduction of needles in the body (for biopsies, punctures, brachytherapy, etc.).

**82.8. Conclusion**

In this chapter, we have introduced the main motivations for Computer-Assisted Medical Interventions and presented the place of medical robots in this general paradigm. Different generations of robots have been proposed and evaluated worldwide, ranging from systems inspired from industrial automation to more specific clinical robots. In parallel to those evolutions in terms of robot architecture and application areas, we have shown that the control modes also evolved in different directions: giving a larger place to the operator through real cooperation or closing the control loop on real-time imaging data for tracking mobile and deformable targets. The future is probably in the merging of such type of controls; for instance the robot might handle synchronization with a moving organ while the operator would control the fine motions with respect to the stabilized target.

As regards industrial products resulting from this domain, the evolution has been very similar. The future will tell if those new design choices will result in a much larger spreading of medical robots. [49] reports 39000 service robots for professional use installed worldwide up to the end of 2006 among which 9% (about 3500 devices) would be medical robots. However, this report does not tell precisely what is included in this category (integration of haptic devices? integration of mechanical localizers? other?). From the numbers mentioned in table 1, our estimate of installed robots in 2008 would be closer to 2500 which is probably less than half the number of installed navigation workstations. Increasing this ratio requires careful selection of applications with significant added value and the development of user-friendly cost-effective systems. Intra-body highly integrated mechatronic devices potentially open the range of applications in a dramatic way. This domain has still to be largely explored.

**Medical glossary**
- **A**neurysm: a local hernia on a blood vessel potentially resulting in vessel rupture and internal haemorrhage.
- Autoclavable: which can be cleaned in the autoclave (pressured vapour sterilization with temperatures greater than 100°C)
- Arthroplasty: plastic surgery of the joints involved for instance in joint replacement (e.g. hip, knee).
- **B**iopsy: the action of taking samples of a tissue with a needle for further analysis.
- Brachytherapy: the introduction of radioactive seeds into an organ for tumour destruction.
- **C**atheter: a flexible tube introduced into the body, typically blood vessels, for instance to inject drugs or to place dilatation devices. Generally, the end of the catheter can be slightly curved by the physician from the outside of the body.
- Colonoscopy: the endoscopic examination of the colon.
- CT (computed tomography): 3D imaging from X-ray acquisition; enables good visualization of bony structures and air cavities.
- **D**RR (Digitally Reconstructed Radiograph): a synthetic X-ray projection image computed from a CT volume being given the position of a virtual source and virtual image plane.
- **E**ndoscopic surgery: surgery involving a minimal access to the body through natural cavities or incisions and a visualization of the internal organs using rigid or flexible optical sensors (the endoscope).
- EC marking: the certification from the European Community necessary for marketing a products of any type in the Economic European Community; it insures that the product complies with the European regulations in terms of safety, health, environment, etc.
- FDA: Food and Drug Administration; the American administration in charge of controlling the safety and efficacy of health-related products (see http://www.fda.gov )
- **I**nterventional radiology: a category of therapeutic or diagnostic procedures executed under imaging control.
- **M**RI (magnetic resonance imaging): 3D imaging from magnetic resonance of hydrogen protons in the body; enables detailed visualization of soft tissues.
- **N**eurosurgery: brain or spine surgery.
- **O**R: operating room also called operating theatre.
- Orthopaedics: a surgical specialty dealing with skeleton bones and joints.
- **P**ET (Positron Emission Tomography): a functional imaging modality. A radioactive marker is associated to a metabolically active molecule and injected into the body of the patient; the metabolic activity is traceable thanks to the radioactive marker and can be reconstructed in 3D thanks to tomography techniques similar to the one used for CT.
- Puncture: the action of inserting a linear tool (a needle or an electrode for instance) into the body.
- Pre-operative, intra-operative, post-operative: before, during, after the intervention.

- **P**rostatectomy: the surgical removal of the prostate in case of cancer; laparoscopic prostatectomy is the minimally invasive version of this surgery.
- **R**adiopaque: visible on X-ray images.
- **R**adiotherapy: the destruction of pathologic tissues (mostly tumours) by ionizing particles.
- **S**tereotactic neurosurgery: a minimally invasive access to the brain requiring a very accurate localization of intracranial structures.
- **S**tereotactic frame: a mechanical device for perfect immobilization of a patient's skull; also used for transferring the surgical plan and for guiding the surgical tool.
- **U**ltrasonic examination: an imaging modality (2D or 3D) based on the propagation of ultrasounds in the body; it visualizes tissue interfaces.